\pdfoutput=1
\documentclass{article} %
\usepackage{iclr2027_conference,times}

\usepackage{amsmath,amsfonts,bm}

\def\eqref#1{equation~\ref{#1}}

\def\1{\bm{1}}

\DeclareMathAlphabet{\mathsfit}{\encodingdefault}{\sfdefault}{m}{sl}
\SetMathAlphabet{\mathsfit}{bold}{\encodingdefault}{\sfdefault}{bx}{n}

\usepackage{hyperref}
\usepackage{url}
\usepackage[utf8]{inputenc} %
\usepackage[T1]{fontenc}    %
\usepackage{hyperref}       %
\usepackage{url}            %
\usepackage{booktabs}       %
\usepackage{amsfonts}       %
\usepackage{nicefrac}       %
\usepackage{microtype}      %
\usepackage{xcolor}         %
\usepackage{amsmath}
\usepackage{fix-cm}
\usepackage{wrapfig}
\usepackage{graphicx}
\usepackage{cleveref}
\usepackage{multirow}
\usepackage[table]{xcolor}
\usepackage{caption}
\usepackage{algorithm}
\usepackage{algpseudocode}
\usepackage{tabularx}
\usepackage{placeins}
\definecolor{refgray}{RGB}{235,242,250}
\definecolor{closedgray}{gray}{0.88}
\usepackage{xcolor}
\newcommand{\reftext}[1]{\textcolor[gray]{0.6}{#1}}
\newcommand{\closedtext}[1]{\textcolor[RGB]{150,170,190}{#1}}
\usepackage{tcolorbox}
\tcbuselibrary{skins}
\newtcolorbox[auto counter]{takeaway}{
    enhanced,
    colback=gray!5,
    colframe=black!50,
    boxrule=0.6pt,
    arc=2pt,
    left=6pt,
    right=6pt,
    top=4pt,
    bottom=0.5pt,
    before skip=8pt,
    after skip=6pt,
    title={Takeaway~\thetcbcounter},
    coltitle=black,
    fonttitle=\bfseries,
    attach boxed title to top left={
        xshift=6pt,
        yshift=-5pt
    },
    boxed title style={
        colback=gray!15,
        colframe=black!50,
        boxrule=0.6pt,
        arc=2pt,
        left=4pt,
        right=4pt,
        top=0pt,
        bottom=0pt
    }
}
\newtcolorbox{keyquestion}{
    enhanced,
    colback=gray!5,
    colframe=black!50,
    boxrule=0.6pt,
    arc=2pt,
    left=6pt,
    right=6pt,
    top=7pt,
    bottom=5pt,
    before skip=8pt,
    after skip=6pt,
    title={Research Question},
    coltitle=black,
    fonttitle=\bfseries,
    attach boxed title to top left={
        xshift=6pt,
        yshift=-3pt
    },
    boxed title style={
        colback=gray!15,
        colframe=black!50,
        boxrule=0.6pt,
        arc=2pt,
        left=4pt,
        right=4pt,
        top=2pt,
        bottom=2pt
    }
}

\title{Soft Spatial Reasoning}

\author{Rafi Ibn Sultan$^{1}$  \quad Md. Sajid Alam Chowdhury$^{1}$ \quad Saleh Zare Zade$^{1}$ \quad \\
\textbf{Chengyin Li$^{2}$} \quad \textbf{Prashant Khanduri$^{1}$} \quad \textbf{Marco Brocanelli$^{3}$} \quad \textbf{Dongxiao Zhu}$^{\textbf{1,4}}$\\
$^{1}$\small Department of Computer Science, Wayne State University \quad\\
$^{2}$\small Department of Radiation Oncology, Henry Ford Health \\
$^{3}$\small Department of Electrical and Computer Engineering, The Ohio State University\quad
\\$^{4}$\small Institute for AI and Data Science, Wayne State University\\
}

\iclrfinalcopy %
\begin{document}

\maketitle
\lhead{Preprint}

\begin{abstract}
Large Vision-Language Models (LVLMs) commonly perform spatial reasoning through chain-of-thought (CoT), encoding intermediate reasoning as autoregressive sequences of discrete language tokens. Such \emph{hard thinking} requires committing to a single token at each step, even when the correct spatial interpretation remains uncertain. This early commitment constitutes \emph{premature discretization}: an incorrect token selection can propagate errors through subsequent reasoning. We propose \textbf{Soft Spatial Reasoning}, a post-training framework that
introduces \emph{soft thinking} for spatial tasks in LVLMs. At each intermediate reasoning step, the LVLM forms a continuous soft state by mixing token embeddings rather than selecting a single token, allowing multiple candidate continuations to influence the next step.
The appropriate degree of softness, however, can vary across reasoning steps: retaining multiple candidates may preserve a useful spatial interpretation, but if those candidates imply conflicting spatial relations, mixing them may
interfere with subsequent reasoning. At the core of Soft Spatial Reasoning is \textbf{AdaptSoft}, a controller that uses the current hidden state and predictive uncertainty to adapt the degree of softness at each reasoning step. To train AdaptSoft, we introduce a gradient-alignment learning objective that provides a step-specific learning signal for softness control without intermediate reasoning supervision. Across diverse spatial benchmarks, Soft Spatial Reasoning outperforms hard and fixed-soft CoT baselines using the same backbone, as well as a range of existing LVLMs. The
source code is available at \url{https://github.com/rafiibnsultan/Soft_Spatial_Reasoning}.
\end{abstract}

\section{Introduction}
\label{sec:intro}

\begin{wrapfigure}{r}{0.55\textwidth}
\vspace{-5mm}
\includegraphics[width=\linewidth]{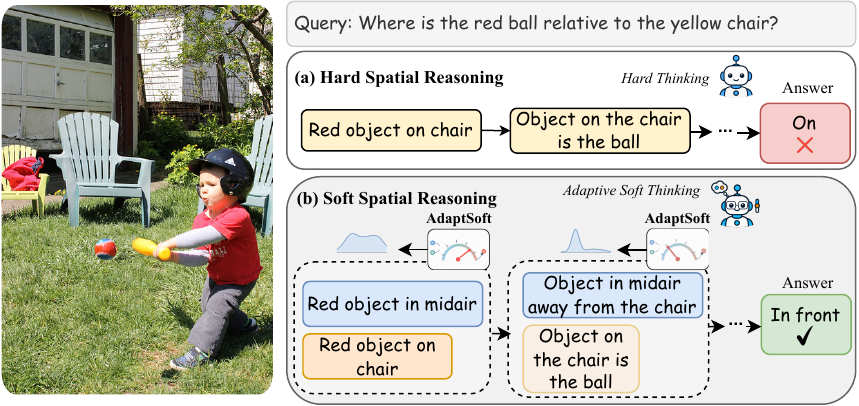}
\caption{\small Hard versus adaptive soft thinking for spatial reasoning. (a) Hard thinking commits to one token at each CoT step, allowing early errors to propagate. (b) Our \textbf{Soft Spatial Reasoning} forms continuous intermediate states by mixing token embeddings, carrying information from multiple candidate continuations through the CoT. \textbf{AdaptSoft} adjusts the mixture's softness at each step based on the current reasoning state and predictive uncertainty.}
\vspace{-5mm}
\label{fig:figure1}
\end{wrapfigure}
Large Vision-Language Models (LVLMs)~\cite{alayrac2022flamingo,li2023blip,liu2023visual,dai2023instructblip,bai2023qwen} have demonstrated capabilities across a range of tasks involving visual understanding~\cite{yin2024survey,li2024survey} and spatial reasoning~\cite{liu2026spatial,stogiannidis2025mind}. To support complex reasoning, these models commonly use chain-of-thought (CoT)~\cite{wei2022chain}, generating intermediate reasoning steps before producing a final answer. Standard CoT expresses these steps as an autoregressive sequence of discrete language tokens, a process we refer to as \emph{hard thinking}. As an alternative to this discrete formulation, recent work on Large Language Models (LLMs)~\cite{hao2024training,zhang2026soft,zheng2025soft} has explored \emph{soft thinking}, in which continuous intermediate states carry information from multiple candidate continuations~\cite{zhu2026reasoning,zhang2026seeing} while the final answer is expressed in language.

LVLMs, however, typically use \emph{hard thinking} for spatial reasoning tasks~\cite{li2025spatialladder,gholami2025spatial,wang2025svqa}. During the CoT, several spatial interpretations may still appear plausible to the model~\cite{zhang2025do}. Consider the question in \Cref{fig:figure1}a, which asks where the red ball is relative to the yellow chair. Before resolving which red object is the ball, the model may identify
the object on the chair as the ball in its CoT, even though the object
in midair remains a plausible candidate. Subsequent reasoning may then treat the object on the chair as the ball, producing the answer \texttt{on} rather than \texttt{in front}. This illustrates \emph{premature discretization}: committing to a discrete continuation before resolving the spatial interpretation can allow an early error to propagate through the remaining CoT~\cite{hao2024training,hu2026colt}. Human cognition offers a useful contrast: people can think through a spatial problem before reaching an answer, without putting every thought into words~\cite{quiroga2005invariant,fedorenko2016language}.

Reasoning without verbalizing every intermediate step is also emerging in LVLMs. Some approaches maintain and refine continuous visual tokens alongside the CoT~\cite{yang2026machine,li2026latent,li2026latentimplicit}. In these approaches, softness lies in the visual representations, while the CoT still advances through discrete token selections and may commit prematurely to one interpretation. Other approaches extend soft thinking to the CoT itself~\cite{hu2026colt,jeon2026vision,wang2026forest}, forming soft states that carry information from multiple candidate continuations and may delay commitment to a single reasoning path. However, the benefit of preserving these alternatives depends on what they represent. In spatial reasoning, alternative interpretations of a scene can imply different spatial relations. During the CoT, a soft state may keep a useful interpretation available at one step; at another, mixing interpretations that imply conflicting relations may interfere with subsequent reasoning. A fixed degree of softness may not suit both situations (\Cref{fig:figure1}b).

\begin{keyquestion}
\emph{How should LVLMs dynamically control representation softness during spatial reasoning?}
\end{keyquestion}

To address this question, we propose \textbf{Soft Spatial Reasoning}
(\Cref{fig:figure1}b), a post-training framework that introduces
\emph{adaptive soft thinking} for spatial reasoning in LVLMs using soft
states at each intermediate CoT step.
Central to the framework is \textbf{AdaptSoft}, a controller that adjusts the softness of the token-embedding mixtures forming these states. AdaptSoft uses the LVLM's hidden state at each step to account for the
ongoing reasoning and predictive uncertainty to gauge the model's
confidence in its next continuation.
By varying softness, it adjusts the distribution of weights across
candidate continuations within each soft state, with the aim of guiding
subsequent reasoning toward a correct final answer.
We further introduce a \textbf{gradient-alignment objective} that measures
agreement between each step's contribution to the LVLM policy gradient
and a reference gradient.
This provides step-level credit alongside task-level rewards for jointly
post-training AdaptSoft and the LVLM, without requiring intermediate
reasoning annotations.

Our contributions are fourfold:
\begin{itemize}
\item We introduce \textbf{Soft Spatial Reasoning}, to our knowledge the first framework dedicated to soft thinking for spatial reasoning in LVLMs.

\item We design \textbf{AdaptSoft} to control softness at each reasoning step using the current reasoning state and predictive uncertainty.

\item We develop a \textbf{gradient-alignment objective} that provides step-specific credit without external evaluators or intermediate reasoning annotations.

\item Extensive experiments covering \textbf{19 spatial reasoning categories} demonstrate overall improvements over strong spatial reasoning baselines.
\end{itemize}

\section{Related Works}
\label{sec:related_works}

\paragraph{Hard Spatial Reasoning in LVLMs.}
Several approaches support spatial reasoning by providing LVLMs with explicit scene geometry and object relationships~\cite{cheng2024spatialrgpt,ma2025spatialllm,cai2025depthlm,liu2025ssr,chen2024ll3da,hu2025g,wang2025n3d,cai2025spatialbot,chensd,sultan2026walkgpt,daxberger2503mm,hong20233d,wu2025spatial,xu2026s,zhao2025spacemind}. Others organize spatial evidence into intermediate representations for grounding and reasoning across perspectives~\cite{bigverdi2025perception,wan2025eaglevision,ning2025enhancing,gholami2025spatial,lee2025perspective,yang2025thinking,chen2026think,zhou2026learning}. Post-training methods optimize grounded reasoning and spatial predictions through supervision or task-specific rewards~\cite{kancheti2026faithful,xu2025visual,zheng2025deepeyes,ma2026thinking,chen2025sifthinker,batra2025spatialthinker,wang2025svqa,wu2025reinforcing,sarch2025grounded,zhao2025embodied,li2025spatialladder,chen2026spacetools,li2026star}, whereas training-free methods guide inference through spatial prompting or interventions on visual processing~\cite{liao2024reasoning,ma2024spatialpin,mitra2024compositional,chen2025spatial,yangrasp}. These methods strengthen spatial evidence and its use, but premature commitment to a single continuation can remain in discrete CoT.

\paragraph{Continuous and Soft Thinking.}
Recent work explores continuous intermediate states in place of discrete CoT generation in LLMs~\cite{hao2024training,wei2026sim,zhou2026lepo}. Soft thinking retains multiple candidate continuations through mixtures of token embeddings~\cite{zhang2026soft,butt2026soft,zheng2025soft,wang2025improving}. Other work explores discrete CoT paths~\cite{dang2026temperature,zhou2026look,yu2026thermometer,wei2026taming} or switches between continuous and discrete reasoning~\cite{shi2026swireasoning,xu2026thinkrouter,xu2026thinking}. In LVLMs, continuous representations support multimodal reasoning, including soft thinking within the CoT~\cite{hu2026colt,sun2026latent,chen2026reasoning,shen2025efficient,pham2026multimodal,ma2025cocova,jeon2026vision,wang2026forest,ray2026mull,huang2026dlwm}, while complementary approaches construct or refine visual representations for subsequent reasoning~\cite{yang2026machine,li2026latent,li2026latentimplicit}. These approaches do not learn step-specific softness for token-embedding mixtures. Soft Spatial Reasoning uses the hidden state and predictive uncertainty to adjust softness, with gradient alignment guiding training.

\section{Method}
\label{sec:method}

\textbf{Soft Spatial Reasoning} (\Cref{fig:figure2}a) is a post-training framework that enables adaptive soft thinking for spatial tasks in LVLMs through continuous intermediate states and discrete final answers. Its controller, \textbf{AdaptSoft} (\Cref{fig:figure2}b), determines step-specific softness from the current hidden state and predictive uncertainty. \textbf{Gradient-Alignment Learning} (\Cref{fig:figure3}) trains this controller by deriving step-specific credit from alignment between each step's policy gradient and a reference gradient.

\subsection{Problem Setup}
\label{subsec:problem_formulation}

Given an image $I$ and a spatial reasoning query $q$, an LVLM policy
$\pi_\theta$ generates a rollout $o=(C,a)$ consisting of a CoT $C$ and
a discrete final answer $a$. Hard thinking represents $C$ as a sequence of discrete language tokens,
each conditioning subsequent predictions. Soft thinking instead represents $C$ as continuous states (\Cref{subsec:soft_cot}).

\begin{figure*}[t]
\centering
\includegraphics[width=0.9\textwidth]{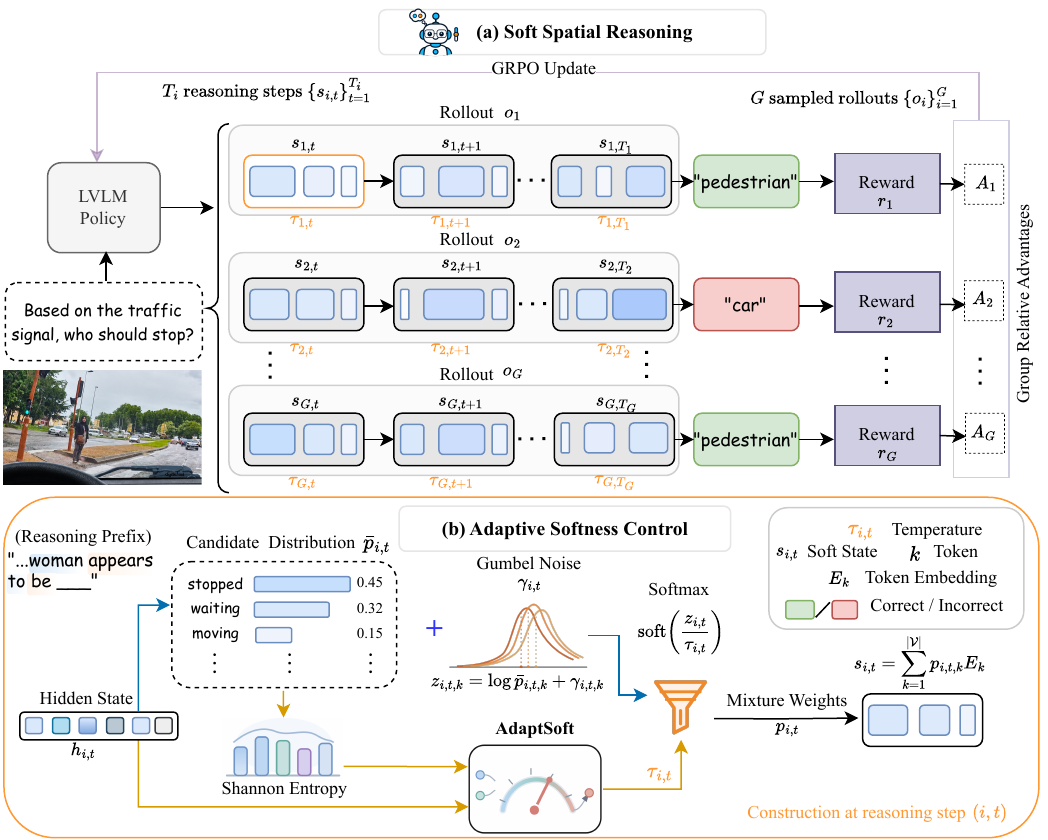}
\caption{\small
Overview of \textbf{Soft Spatial Reasoning}.
(a) The LVLM policy is post-trained to carry multiple candidate continuations through soft states during reasoning, then generate the final answer as discrete tokens. Rollout rewards yield group-relative advantages for the GRPO update.
(b) At each reasoning step, \textbf{AdaptSoft} sets the mixture temperature from the current hidden state and candidate-distribution entropy. The resulting mixture weights combine token embeddings into the next soft state.
}

\label{fig:figure2}
\vspace{-10pt}
\end{figure*}

The LVLM policy is post-trained using Group Relative Policy Optimization
(GRPO)~\cite{shao2024deepseekmath}, which uses relative rewards within groups
of sampled rollouts without requiring ground-truth CoT supervision.
As illustrated in \Cref{fig:figure2}a, for each image--query pair $(I,q)$, the rollout policy
$\pi_{\theta_{\mathrm{old}}}$ samples a group of $G$ rollouts:
\begin{equation}
o_i
\sim
\pi_{\theta_{\mathrm{old}}}(\cdot\mid I,q),
\qquad
i=1,\ldots,G.
\label{eq:group_rollout}
\end{equation}
Each rollout receives an answer reward $R_{\mathrm{ans}}^{(i)}$ and a format reward
$R_{\mathrm{fmt}}^{(i)}$. The answer reward assigns full credit ($1$) when the final answer is correct
and no credit ($0$) otherwise, while the format reward evaluates compliance
with the required reasoning-and-answer structure. Using reward weights
$w_{\mathrm{ans}}$ and $w_{\mathrm{fmt}}$, and a small constant $\epsilon$ for
numerical stability, we define the task reward and its group-normalized
advantage as
\begin{equation}
r_i
=
w_{\mathrm{ans}}R_{\mathrm{ans}}^{(i)}
+
w_{\mathrm{fmt}}R_{\mathrm{fmt}}^{(i)},
\qquad
A_i^{\mathrm{task}}
=
\frac{
r_i-\operatorname{mean}(\{r_j\}_{j=1}^{G})
}{
\operatorname{std}(\{r_j\}_{j=1}^{G})+\epsilon
}.
\label{eq:task_reward_and_advantage}
\end{equation}

\subsection{Soft Chain-of-Thought (CoT)}
\label{subsec:soft_cot}

Within each rollout $o_i$, the intermediate CoT is represented by a sequence
of $T_i$ soft states $\{s_{i,t}\}_{t=1}^{T_i}$, formed by feeding continuous
mixtures of token embeddings back into the language backbone at successive
reasoning steps (\Cref{fig:figure2}a)~\cite{zhang2026soft,butt2026soft}. After the reasoning steps, the model switches to discrete token generation for the final answer.

\paragraph{Stochastic Soft Rollout.}
To produce different soft CoTs within the rollout group, independent Gumbel perturbations are applied to the
vocabulary log-probabilities at each reasoning
step~\cite{zheng2025soft,wu2026llms}, as illustrated in
\Cref{fig:figure2}b. Specifically, conditioned on $(I,q)$ and the preceding states $s_{i,<t}$, the
rollout policy produces the unperturbed distribution $\bar{p}_{i,t}$. For each
token $k$, an independent Gumbel sample $\gamma_{i,t,k}$ is then added to
$\log \bar{p}_{i,t,k}$, yielding the perturbed score $z_{i,t,k}$:
\begin{equation}
\begin{gathered}
\bar{p}_{i,t}
=
\pi_{\theta_{\mathrm{old}}}
\left(
\cdot\mid I,q,s_{i,<t}
\right),
\qquad
\gamma_{i,t,k}
\overset{\mathrm{i.i.d.}}{\sim}
\operatorname{Gumbel}(0,1),
\\
z_{i,t,k}
=
\log \bar{p}_{i,t,k}+\gamma_{i,t,k}.
\end{gathered}
\label{eq:gumbel_perturbation}
\end{equation}
A temperature-controlled softmax (\Cref{fig:figure2}b) converts the perturbed scores into
token-mixture weights $p_{i,t}$, whose weighted combination of token embeddings
forms the soft state $s_{i,t}$:
\begin{equation}
p_{i,t}
=
\operatorname{softmax}\!\left(
\frac{z_{i,t}}{\tau_{i,t}}
\right),
\qquad
s_{i,t}
=
\sum_{k=1}^{|\mathcal{V}|}
p_{i,t,k}E_k,
\label{eq:gumbel_soft_state}
\end{equation}
where $\mathcal{V}$ is the language vocabulary. At each step, the softmax
is applied to its top-$\mathcal{K}$ candidate tokens, with $p_{i,t,k}=0$
for all other tokens. Here, $k$ indexes vocabulary tokens, and
$E_k\in\mathbb{R}^{d}$ is the embedding of token $k$. The temperature $\tau_{i,t}>0$ controls mixture concentration: lower
values move $s_{i,t}$ toward the highest-scoring token's embedding,
while higher values spread weight across token embeddings.

\paragraph{Gumbel-Reparameterized Likelihood.}
GRPO requires a likelihood ratio between the current and rollout policies
at each reasoning step. Unlike a discrete CoT step, a soft state $s_{i,t}$
combines token embeddings without selecting an individual token, so the
standard sampled-token likelihood does not apply. The perturbed score vector $z_{i,t}$ is the sampled variable that,
given $\tau_{i,t}$, deterministically specifies $s_{i,t}$
(\Cref{eq:gumbel_perturbation,eq:gumbel_soft_state}). The likelihood ratio
is therefore defined over $z_{i,t}$, evaluating the same recorded score
vector under both policies.

During the policy update, the sampled scores and corresponding temperatures
from the preceding steps reconstruct $s_{i,<t}$, ensuring that both policies
are evaluated on the same reasoning prefix.
Conditioned on $(I,q,s_{i,<t})$, the current policy produces
$\bar{p}_{i,t}^{\,\theta}$. For each recorded score $z_{i,t,k}$, subtracting the current token
log-probability yields the Gumbel noise value implied by the current
policy. Evaluating the implied noise values under independent standard Gumbel
densities gives the conditional joint log-density of $z_{i,t}$, with
conditioning on $(I,q,s_{i,<t})$ omitted for compactness:
\begin{equation}
\begin{gathered}
\bar{p}_{i,t}^{\,\theta}
=
\pi_\theta(\cdot\mid I,q,s_{i,<t}),
\qquad
\widetilde{\gamma}_{i,t,k}^{\,\theta}
=
z_{i,t,k}-\log \bar{p}_{i,t,k}^{\,\theta},
\\
\log P_\theta(z_{i,t})
=
\sum_{k=1}^{|\mathcal{V}|}
\left[
-\widetilde{\gamma}_{i,t,k}^{\,\theta}
-
\exp\!\left(-\widetilde{\gamma}_{i,t,k}^{\,\theta}\right)
\right].
\end{gathered}
\label{eq:current_gumbel_likelihood}
\end{equation}
Evaluating the same scores under the rollout policy gives
$P_{\theta_{\mathrm{old}}}(z_{i,t})$, yielding the soft-step GRPO ratio:
\begin{equation}
\rho_{i,t}^{\mathrm{soft}}(\theta)
=
\exp\!\left(
\log P_\theta(z_{i,t})
-
\log P_{\theta_{\mathrm{old}}}(z_{i,t})
\right).
\label{eq:soft_likelihood_ratio}
\end{equation}

As detailed in Section~\ref{subsec:adaptsoft}, the rollout temperatures are
reintroduced through AdaptSoft when reconstructing the soft states, allowing
gradients from subsequent reasoning steps to propagate to its parameters. A derivation of the density and likelihood ratio is provided in Appendix
\Cref{app:gumbel_likelihood}.

\subsection{AdaptSoft: Adaptive Softness Control}
\label{subsec:adaptsoft}

\paragraph{Step-Specific Softness Controller.}
At reasoning step $t$ of rollout $o_i$, let $h_{i,t}\in\mathbb{R}^{d}$
denote the final-layer hidden state and $\bar{p}_{i,t}$ the unperturbed
next-token distribution. The hidden state summarizes the accumulated reasoning
context, while the entropy of $\bar{p}_{i,t}$ measures predictive uncertainty
among candidate continuations. The entropy is standardized as
\begin{equation}
\widehat{H}_{i,t}
=
\frac{H(\bar{p}_{i,t})-\mu_H}
{\sigma_H+\epsilon_H},
\label{eq:token_uncertainty}
\end{equation}
where $H(\cdot)$ denotes Shannon entropy. The mean $\mu_H$ and standard
deviation $\sigma_H$ are estimated once from the next-token entropies of the
initial policy and remain fixed throughout post-training. The constant
$\epsilon_H>0$ prevents division by zero.

Before entering the controller, $h_{i,t}$ is layer-normalized and projected
to $d_p<d$ dimensions using a fixed random matrix
$\mathbf{P}\in\mathbb{R}^{d_p\times d}$. This compact representation reduces
the rollout storage required for controller training. The projected state is
concatenated with $\widehat{H}_{i,t}$ and passed to an MLP $f_\phi$, whose
scalar output parameterizes the temperature:
\begin{equation}
u_{i,t}
=
f_\phi\!\left(
\left[
\mathbf{P}\,\operatorname{LN}(h_{i,t})
\;\middle\Vert\;
\widehat{H}_{i,t}
\right]
\right),
\qquad
\tau_{i,t}
=
\tau_0+\Delta\tanh(u_{i,t}),
\label{eq:adaptive_temperature}
\end{equation}
where $\operatorname{LN}$ denotes layer normalization, $\Vert$ denotes
concatenation, and $\phi$ contains the learnable controller parameters.
Since $\tanh(u_{i,t})\in(-1,1)$, $\tau_0$ sets the base temperature and
$\Delta$ sets its maximum step-specific deviation, yielding
$\tau_{i,t}\in(\tau_0-\Delta,\tau_0+\Delta)$. We require
$0<\Delta<\tau_0$ to keep all temperatures above zero.

Within this range, AdaptSoft uses the current reasoning context and
predictive uncertainty to adjust the softness of each mixture. As described in \Cref{subsec:soft_cot}, the sampled soft states are
reconstructed during the policy update from the recorded scores and rollout
temperatures. To train
AdaptSoft without altering these sampled states, the controller inputs and
outputs are retained during rollout. A stop-gradient construction uses the
recorded temperature in the forward computation while passing gradients
through its recomputed value to $\phi$; details are provided in Appendix
\Cref{app:adaptsoft_stopgrad}.

\subsection{Post-Training Optimization}
\label{subsec:gradient_alignment_learning}

The LVLM policy is optimized with GRPO, while AdaptSoft is trained through
step-specific gradient alignment.

\paragraph{GRPO for the LVLM Policy.} For each sampled step,
$\rho_{i,t}(\theta)$ compares its likelihood under the current and rollout
policies. Soft reasoning steps use the density ratio
$\rho_{i,t}^{\mathrm{soft}}(\theta)$ in
Equation~\ref{eq:soft_likelihood_ratio}, while discrete final-answer tokens
use the token-probability ratio. Each ratio is weighted by the shared
rollout advantage $A_i^{\mathrm{task}}$, encouraging steps from
higher-reward rollouts and discouraging those from lower-reward rollouts.
The clipped surrogate limits the incentive for large ratio changes,
while KL regularization over the LVLM's next-token distributions
penalizes deviation from $\pi_{\mathrm{ref}}$.
Averaging over steps and rollouts gives
\begin{equation}
\begin{aligned}
\mathcal{J}(\theta)
={}&
\mathbb{E}_{\{o_i\}\sim\pi_{\theta_{\mathrm{old}}}}
\Bigg[
\frac{1}{G}\sum_{i=1}^{G}
\frac{1}{|o_i|}\sum_{t=1}^{|o_i|}
\Big(
\min\big\{
\rho_{i,t}(\theta)A_i^{\mathrm{task}},
\\[-2pt]
&\qquad
\operatorname{clip}\big(
\rho_{i,t}(\theta),1-\delta,1+\delta
\big)A_i^{\mathrm{task}}
\big\}
-\eta D_{\mathrm{KL}}
\big(\pi_\theta\,\|\,\pi_{\mathrm{ref}}\big)
\Big)
\Bigg].
\end{aligned}
\label{eq:grpo_objective}
\end{equation}
Here, $|o_i|$ counts soft reasoning steps and discrete final-answer tokens,
$\delta$ is the clipping threshold, and $\eta$ weights the KL penalty
against the reference LVLM policy $\pi_{\mathrm{ref}}$. The KL distributions
are conditioned on $(I,q)$ and the preceding steps. Optimization minimizes
$\mathcal{L}_{\mathrm{GRPO}}(\theta)=-\mathcal{J}(\theta)$.

\begin{wrapfigure}{r}{0.48\textwidth}
\vspace{-5mm}
\includegraphics[width=\linewidth]{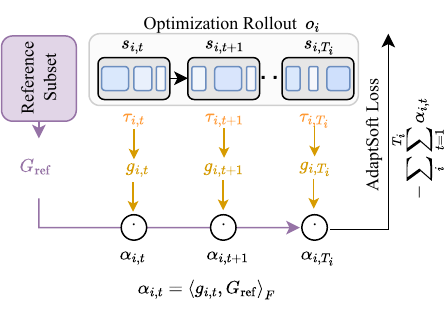}
\vspace{-8mm}
\caption{\small \textbf{Gradient-Alignment Learning.}
At each soft reasoning step $t$, the temperature $\tau_{i,t}$ controls the
state $s_{i,t}$, whose contribution $g_{i,t}$ to the LVLM output-layer
gradient is compared with a reference gradient $G_{\mathrm{ref}}$ computed
from a disjoint subset of rollouts. The resulting alignment scores
$\alpha_{i,t}$ are aggregated to form the AdaptSoft loss.}
\vspace{-5mm}
\label{fig:figure3}
\end{wrapfigure}
\paragraph{Step-Specific Gradient Alignment for AdaptSoft.}
To train AdaptSoft, we introduce a gradient-alignment objective that
provides a step-specific learning signal for each predicted temperature.
The shared rollout-level advantage in GRPO does not directly distinguish
how softness should vary across reasoning steps. The proposed objective
provides this signal by comparing each step's contribution to the LVLM
policy gradient with a reference gradient (\Cref{fig:figure3}). Each batch of sampled rollouts is
divided into disjoint reference and optimization subsets. The reference subset
provides a gradient $G_{\mathrm{ref}}$, while each soft reasoning step
$t$ of rollout $o_i$ in the optimization subset contributes a gradient
$g_{i,t}$. Both gradients are computed with respect to the LVLM
output-layer weights $W$.

The alignment score $\alpha_{i,t}$ is defined as the Frobenius inner
product of these gradients. The AdaptSoft loss
$\mathcal{L}_{\mathrm{AS}}(\phi)$ is the negative sum of the alignment
scores across soft reasoning steps in the optimization subset:
\begin{equation}
\begin{gathered}
G_{\mathrm{ref}}
=
\nabla_W\mathcal{L}_{\mathrm{GRPO}}^{\mathrm{ref}},
\qquad
g_{i,t}
=
\nabla_W\mathcal{L}_{\mathrm{GRPO}}^{(i,t)},
\\
\alpha_{i,t}
=
\left\langle g_{i,t},G_{\mathrm{ref}}\right\rangle_F,
\qquad
\mathcal{L}_{\mathrm{AS}}(\phi)
=
-\sum_i\sum_{t=1}^{T_i}\alpha_{i,t},
\end{gathered}
\label{eq:step_alignment}
\end{equation}
where $\mathcal{L}_{\mathrm{GRPO}}^{\mathrm{ref}}$ is the GRPO loss on
the reference subset, $\mathcal{L}_{\mathrm{GRPO}}^{(i,t)}$ is the
contribution of step $t$ in rollout $o_i$ to the GRPO loss, and $i$
ranges over the optimization subset. For a small update
$W'=W-\lambda g_{i,t}$ with step size $\lambda>0$, the first-order
change in the reference loss is $-\lambda\alpha_{i,t}$. Minimizing
$\mathcal{L}_{\mathrm{AS}}$ thus favors temperatures whose resulting
step-specific gradients align with the reference gradient.

The gradients $g_{i,t}$ are computed through the reconstructed soft
states and remain differentiable with respect to the temperatures.
Holding $G_{\mathrm{ref}}$ and the GRPO logit residuals fixed,
differentiation provides a first-order alignment signal to
temperatures through subsequent reasoning steps. To emphasize differences across reasoning steps, the mean temperature gradient is subtracted within each rollout:
\begin{equation}
\widetilde{g}_{i,t}^{\,\tau}
=
\frac{\partial\mathcal{L}_{\mathrm{AS}}}{\partial\tau_{i,t}}
-
\frac{1}{T_i}
\sum_{t'=1}^{T_i}
\frac{\partial\mathcal{L}_{\mathrm{AS}}}{\partial\tau_{i,t'}}.
\label{eq:adaptsoft_learning}
\end{equation}
The centered gradients satisfy
$\sum_{t=1}^{T_i}\widetilde{g}_{i,t}^{\,\tau}=0$.
Alignment scores are computed using the outer-product structure of
the output-layer gradients, avoiding a separate
$|\mathcal{V}|\times d$ gradient matrix for each soft reasoning step.

\paragraph{Parameter Updates.}
Each training step updates the LVLM parameters $\theta$ using
$\mathcal{L}_{\mathrm{GRPO}}$ from
Equation~\ref{eq:grpo_objective}. The AdaptSoft parameters $\phi$ are
updated by backpropagating the centered temperature gradients from
Equation~\ref{eq:adaptsoft_learning} through the controller in
Equation~\ref{eq:adaptive_temperature} (detailed in Appendix \Cref{app:adaptsoft_gradient}).

\section{Experiments}
\label{sec:experiments}

\begin{table*}[t]
\centering
\small
\caption{\small OmniSpatial \cite{jia2025omnispatial} results across 10 spatial reasoning task categories.
Soft Spatial Reasoning is compared against proprietary models, general open-source LVLMs, soft thinking models, and
specialized spatial reasoning models. Proprietary models are included as reference
points; the best open-source or specialized result is shown in \textbf{bold}.
Average accuracy is weighted by category sample size.}
\resizebox{\textwidth}{!}{
\begin{tabular}{lccccccccccc}
\toprule
\midrule
\multirow{2}{*}{Method} & \multirow{2}{*}{Average}
& \multicolumn{2}{c}{Dynamic Reasoning}
& \multicolumn{3}{c}{Spatial Interaction}
& \multicolumn{2}{c}{Complex Logic}
& \multicolumn{3}{c}{Perspective Taking} \\
\cmidrule(lr){3-4} \cmidrule(lr){5-7} \cmidrule(lr){8-9} \cmidrule(lr){10-12}
& & \shortstack[c]{Mani-\\pulation} & \shortstack[c]{Motion\\Anal.} & \shortstack[c]{Traffic\\Anal.}
& \shortstack[c]{Loca-\\lization} & \shortstack[c]{Geospa.\\Strategy} & \shortstack[c]{Pattern\\Rec.}
& \shortstack[c]{Geometric\\Reasoning}
& \shortstack[c]{Ego\\Centric} & \shortstack[c]{Allo\\Centric} & \shortstack[c]{Hypo-\\thetical} \\
\hline

\multicolumn{12}{l}{\textbf{Reference Baselines}} \\
\reftext{Random Choice} & \reftext{24.98} & \reftext{24.86} & \reftext{26.30} & \reftext{25.88} & \reftext{23.43} & \reftext{27.27} & \reftext{21.44} & \reftext{24.77} & \reftext{22.55} & \reftext{24.84} & \reftext{25.78} \\
\reftext{Human Evaluation} & \reftext{92.63} & \reftext{94.62} & \reftext{96.07} & \reftext{91.38} & \reftext{95.11} & \reftext{92.15} & \reftext{89.02} & \reftext{85.90} & \reftext{98.53} & \reftext{94.30} & \reftext{90.26} \\
\hline
\multicolumn{12}{l}{\textbf{Proprietary Models}} \\
\closedtext{GPT-4.1-mini} \closedtext{\cite{open2025introducing}} & \closedtext{48.87} & \closedtext{64.32} & \closedtext{56.53} & \closedtext{59.06} & \closedtext{60.19} & \closedtext{56.36} & \closedtext{29.28} & \closedtext{30.19} & \closedtext{72.55} & \closedtext{39.57} & \closedtext{39.28} \\
\closedtext{Gemini-2.5-flash-preview} \closedtext{\cite{team2023gemini}} & \closedtext{52.12} & \closedtext{67.57} & \closedtext{62.72} & \closedtext{68.24} & \closedtext{73.33} & \closedtext{60.91} & \closedtext{38.14} & \closedtext{34.19} & \closedtext{75.49} & \closedtext{35.90} & \closedtext{33.73} \\
\closedtext{o4-mini} \closedtext{\cite{openai2025o3o4mini_system_card}} & \closedtext{52.77} & \closedtext{72.97} & \closedtext{59.83} & \closedtext{60.00} & \closedtext{73.33} & \closedtext{61.82} & \closedtext{34.02} & \closedtext{36.77} & \closedtext{73.53} & \closedtext{40.69} & \closedtext{40.96} \\
\closedtext{Gemini-2.5-flash} \closedtext{\cite{team2023gemini}} & \closedtext{53.16} & \closedtext{70.27} & \closedtext{64.74} & \closedtext{61.18} & \closedtext{72.38} & \closedtext{58.18} & \closedtext{35.05} & \closedtext{36.13} & \closedtext{74.12} & \closedtext{40.96} & \closedtext{32.53} \\
\hline

\multicolumn{12}{l}{\textbf{Open-weights Models}} \\

LLaVA-1.5-7B \cite{liu2024improved}& 34.97 & 54.46 & 31.23 & 35.29 & 36.19 & 33.94 & 29.01 & 24.18 & 55.60 & 34.66 & 36.14 \\
InternVL3-8B \cite{zhu2025internvl3}& 41.60 & 52.43 & 40.87 & 48.94 & 51.05 & 44.77 & 24.95 & 28.63 & 64.20 & 38.62 & 40.96 \\
InternVL3-14B \cite{zhu2025internvl3}& 45.94 & 54.32 & 60.17 & 50.35 & 51.81 & 51.45 & 28.04 & 28.26 & 68.04 & 35.37 & 34.46 \\

Qwen2.5-VL-7B \cite{wang2024qwen2}& 39.18 & 58.38 & 35.09 & 50.12 & 45.33 & 44.00 & 31.13 & 29.42 & 64.51 & 33.19 & 37.35 \\
Gemma-3-12B \cite{gemmateam2025gemma3technicalreport}& 43.71 & 54.05 & 54.91 & 54.12 & 47.62 & 45.45 & 16.49 & 30.32 & 63.73 & 36.70 & 33.73 \\

\hline

\multicolumn{12}{l}{\textbf{Soft Thinking Models}} \\
LVR \cite{li2026latent}& 41.86 & 54.05 & 35.83 & 44.70 & 53.33 & 48.18 & 28.86 & 27.74 & 72.54 & 38.29 & \textbf{50.60} \\
Laser \cite{wang2026forest}& 42.78 & 55.40 & 45.95 & 57.64 & 54.28 & 50.00 & 20.61 & 24.51 & 72.54 & 33.51 & 44.57 \\
LaCoT \cite{sun2026latent}& 45.66 & 59.45 & 53.75& 56.47 & 50.47 & 48.18 & 28.86 & 33.54 & 67.64 & 34.57 & 44.57 \\

\hline

\multicolumn{12}{l}{\textbf{Spatial Reasoning Models}} \\
SpaceMantis-13B \cite{chen2024spatialvlm}& 36.36 & 47.03 & 36.59 & 40.94 & 34.86 & 33.09 & 22.27 & 24.39 & 49.22 & 38.25 & 39.28 \\
SpaceQwen2.5-VL-3B \cite{chen2024spatialvlm}& 40.25 & 58.11 & 39.88 & 41.18 & 40.95 & 40.91 & 29.90 & 25.81 & 63.73 & \textbf{38.83} & 39.76 \\
SpaceThinkerQwen2.5VL-3B \cite{chen2024spatialvlm}& 40.42 & 47.84 & 53.06 & 43.29 & 35.43 & 38.73 & 24.33 & 28.00 & 58.04 & 35.11 & 31.08 \\
VST-RL-7B \cite{yang2025visual}& 41.09 & 56.75 & 43.39 & 44.75 & 46.66 & 42.72 & 25.51 & 28.38 & 72.54 & 32.89 & 43.37 \\
SoFar-Qwen2.5VL-3B \cite{qi2025sofar}& 45.14 & 56.49 & 51.16 & 54.12 & 53.14 & \textbf{52.73} & \textbf{31.75} & 22.88 & 71.60 & 36.56 & 41.69 \\
SpatialLadder-3B \cite{li2025spatialladder}& 40.50 & 59.45 & 39.01 & 50.58 & 48.57 & 42.72 & 26.80 & 23.87 & 71.56 & 35.10 & 39.75 \\

\hline

\multicolumn{12}{l}{\textbf{Backbone and Our Variants}} \\

Qwen3-VL-8B-Thinking (Base)~\cite{bai2025qwen3} & 43.90 & 57.14 & 54.28 & 36.04 & 54.95 & 48.67 & 24.77 & 23.12 & 71.56 & 31.11 & 38.82 \\
Hard Thinking + GRPO & 45.92 & 61.32 & 54.30 & 55.88 & 58.33 & 46.18 & 25.90 & 25.00 & 72.92 & 36.57 & 42.53 \\
Soft Thinking + GRPO & 46.93 & 61.82 & 55.85 & 52.21 & 62.62 & 49.77 & 25.26  & 25.00 & \textbf{76.35} & 36.23  & 45.88 \\
\textbf{Soft Spatial Reasoning (Ours)} &\textbf{49.68} & \textbf{63.51} & \textbf{62.14} & \textbf{57.65}& \textbf{62.86} & 50.00 & 29.90 & \textbf{34.84} & 75.49 & 34.93 & 46.34  \\

\midrule
\bottomrule

\end{tabular}
}
\label{tab:table1}
\end{table*}

\begin{figure*}[t]
\centering
\includegraphics[width=0.9\textwidth]{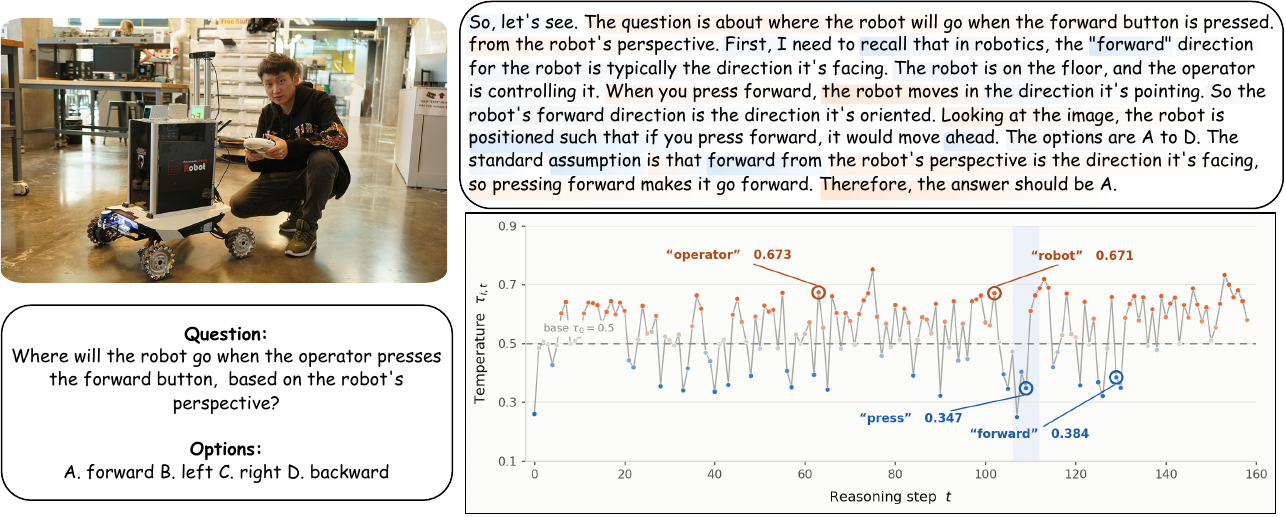}
\caption{\small
Soft Spatial Reasoning on a randomly selected example from OmniSpatial.
AdaptSoft varies softness across the CoT. For readability, greedy
decoding is used only to display the soft states as text. Shading shows
each span's mean temperature (\textcolor{blue}{blue}: lower; \textcolor{orange}{orange}: higher). The plot
below shows temperature at every reasoning step and labels several words with their temperatures.
}

\label{fig:figure4}
\vspace{-5mm}
\end{figure*}

\subsection{Implementation Details}
\label{sec:implementation}

We instantiate Soft Spatial Reasoning with a Qwen3-VL-8B-Thinking
backbone~\cite{bai2025qwen3} and train for two epochs on the OmniSpatial
training split~\cite{jia2025omnispatial}. GRPO uses $G=8$ rollouts per
prompt, a batch size of 64 prompts, and two optimizer updates per batch.
The LVLM policy is optimized with AdamW at a learning rate of $10^{-6}$ and
regularized toward its frozen initialization using a KL coefficient of
$10^{-3}$. AdaptSoft is implemented as a two-layer MLP with projection
dimension $d_p=8$ and optimized with AdamW at a learning rate of $10^{-3}$.
Additional optimization, preprocessing, sequence-length, and system details
are provided in Appendix~\Cref{app:implementation_details}.

\subsection{Baselines}

We compare Soft Spatial Reasoning against open-source LVLMs, including models
specialized in spatial reasoning and soft thinking, and report proprietary
models as reference points. We also train controlled hard- and soft-thinking
baselines using the same backbone and post-training setup. Evaluation covers
three complementary settings. The held-out OmniSpatial test
set~\cite{jia2025omnispatial} evaluates post-training performance across its
spatial-reasoning taxonomy. SpatiaLab~\cite{wasi2026spatialab} measures
zero-shot transfer to realistic, unconstrained scenes from an unseen
benchmark. MindCube~\cite{wang2026mindcubespatialmentalmodeling} evaluates
zero-shot spatial mental modeling from limited views through cognitive
mapping, perspective-taking, and mental simulation. Dataset and evaluation
details are provided in Appendix Section~\ref{app:benchmark_details}.

\subsection{Results}

\begin{table*}[t]
\centering
\small
\caption{\small SpatiaLab~\cite{wasi2026spatialab} results across 6 spatial
reasoning task categories in the zero-shot setting. Proprietary models are included as reference points; the best open-source
or specialized result is shown in \textbf{bold}. Average accuracy is weighted by category sample size.}
\resizebox{\textwidth}{!}{
\begin{tabular}{lccccccc}
\toprule
\midrule
\multirow{2}{*}{Method} & \multirow{2}{*}{Average}
& \multicolumn{6}{c}{Question Categories} \\
\cmidrule(lr){3-8}
& & 3D Geom. & Dep. \& Occu. & Orientation & Relat. Posit. & Size \& Scale & Spati. Navig. \\
\hline

\multicolumn{8}{l}{\textbf{Reference Baselines}} \\
\reftext{Random Choice} & \reftext{25.00} & \reftext{25.00} & \reftext{25.00} & \reftext{25.00} & \reftext{25.00} & \reftext{25.00} & \reftext{25.00} \\
\reftext{Human Baseline} & \reftext{87.57} & \reftext{93.70} & \reftext{74.13} & \reftext{91.58} & \reftext{91.51} & \reftext{88.89} & \reftext{87.76} \\
\hline
\multicolumn{8}{l}{\textbf{Proprietary Models}} \\
\closedtext{GPT-4o-mini} \closedtext{\cite{hurst2024gpt}} & \closedtext{46.50} & \closedtext{47.06} & \closedtext{39.00} & \closedtext{47.03} & \closedtext{47.17} & \closedtext{49.60} & \closedtext{49.79} \\
\closedtext{Gemini-2.5-Flash} \closedtext{\cite{team2023gemini}} & \closedtext{48.29} & \closedtext{44.96} & \closedtext{48.26} & \closedtext{48.02} & \closedtext{56.13} & \closedtext{42.46} & \closedtext{51.05} \\
\closedtext{Mistral Medium 3.1} \closedtext{\cite{mistralai2025medium31}} & \closedtext{47.93} & \closedtext{46.64} & \closedtext{49.81} & \closedtext{47.52} & \closedtext{61.79} & \closedtext{41.67} & \closedtext{41.77} \\

\hline

\multicolumn{8}{l}{\textbf{Open-weights Models}} \\

Qwen2.5-VL-3B-Instruct \cite{wang2024qwen2}& 41.43 & 41.18 & 35.52 & 46.04 & 40.09 & 47.22 & 39.24 \\

InternVL3.5-4B \cite{wang2025internvl3}& 43.29 & 42.86 & 42.86 & 42.08 & 54.72 & 36.51 & 42.19 \\
Gemma-3-4B-it \cite{gemmateam2025gemma3technicalreport}& 40.57 & 43.70 & 34.36 & 46.53 & 45.75 & 37.30 & 37.97 \\
LLaVA-1.5-7B \cite{liu2024improved} & 38.64 & 40.33 & 34.74 & 31.68 & 40.56 & 40.07 & 43.88 \\
Qwen2.5-VL-7B-Instruct \cite{wang2024qwen2}& 41.00 & 42.86 & 37.84 & 42.57 & 46.23 & 42.06 & 35.44 \\

\hline

\multicolumn{8}{l}{\textbf{Soft Thinking Models}} \\
LVR \cite{li2026latent}& 45.78 & \textbf{48.31 }& 43.62 & \textbf{51.98} &49.52& 40.07& 43.03\\
Laser \cite{wang2026forest}& 44.78 & 46.21 & 43.24 & 46.03 & 50.47& 40.07& 43.88\\
LaCoT \cite{sun2026latent}& 45.50 & 46.63 & 40.15 & 47.52 & 53.30& 43.25& 43.88\\

\multicolumn{8}{l}{\textbf{Spatial Reasoning Models}} \\

SpaceOm \cite{chen2024spatialvlm}& 41.36 & 42.44 & 38.61 & 48.02 & 37.74 & 42.86 & 39.24 \\
SpaceThinker-Qwen2.5VL-3B \cite{chen2024spatialvlm}& 40.64 & 40.34 & 37.84 & 47.03 & 38.21 & 43.25 & 37.97 \\
SpaceQwen2.5-VL-3B-Instruct \cite{chen2024spatialvlm}& 40.14 & 31.51 & 35.14 & 37.62 & 37.74 & \textbf{50.79} & 47.26 \\
SpatialLadder-3B \cite{li2025spatialladder}& 35.28 & 39.91 & 34.61 & 39.90 & 38.20 & 25.39 & 33.19\\
\hline

\multicolumn{8}{l}{\textbf{Backbone and Our Variants}} \\

Qwen3-VL-8B-Thinking (Base)~\cite{bai2025qwen3}& 44.78 & 42.01 & 47.10 & 47.02 & 51.88 & 40.47 & 41.35 \\
Hard Thinking + GRPO& 46.21 & 43.70& 47.88 & 47.52 & 54.25& 41.27& 43.88\\
Soft Thinking + GRPO& 46.86 & 44.12 & 48.26 & 48.02 & 55.66& 41.67& 44.73\\
\textbf{Soft Spatial Reasoning (Ours)}& \textbf{48.71}&46.22 & \textbf{49.42} & 48.51& \textbf{58.96}& 42.46& \textbf{48.10}\\
\midrule
\bottomrule

\end{tabular}
}
\label{tab:table2}
\end{table*}

\noindent\textbf{What Adaptive Softness Adds to Spatial Reasoning.}
On OmniSpatial, Soft Spatial Reasoning leads the non-proprietary models
in weighted-average accuracy (\Cref{tab:table1}), surpassing InternVL3-14B,
SoFar, and LaCoT by 3.74, 4.54, and 4.02 percentage points, respectively.
With the same backbone and post-training setup, basic soft thinking
improves on hard thinking by 1.01 points, while adaptive softness adds
another 2.75 points. Geometric Reasoning illustrates this distinction:
hard and basic soft thinking both score 25.00, whereas adaptation reaches
34.84. Motion Analysis also rises from 55.85 to 62.14 over basic soft
thinking. The absence of a geometric gain from basic soft thinking
suggests that preserving alternatives alone may not suffice.
Geometry and motion can involve competing spatial configurations whose
usefulness changes across reasoning steps. Adaptive softness may help
retain useful possibilities while limiting interference from conflicting
relations as the reasoning context changes. Relative to the original
backbone, the full model improves in all ten categories, including a
21.61-point gain in Traffic Analysis.

\Cref{fig:figure4} illustrates how Soft Spatial Reasoning answers a spatial
question about an image. The displayed CoT traces the model's reasoning
about the robot's movement, reaching the correct answer as softness varies
across steps. The text highlights connect this reasoning to the temperature
plot below: ``operator'' and ``robot'' correspond to higher temperatures,
while ``press'' and ``forward'' correspond to lower temperatures. These
adjustments let broader candidate continuations contribute at some steps
and concentrate their influence at others, showing how adaptive softness
operates throughout reasoning. Further analysis of temperature variation
appears in \Cref{app:additional_analysis}.

\noindent\textbf{What Adaptive Softness Adds in Zero-Shot Transfer.}
On the unseen SpatiaLab benchmark (\Cref{tab:table2}), Soft Spatial
Reasoning achieves 48.71, exceeding the strongest prior open-weights
model, LVR, by 2.93 percentage points. The matched-backbone comparison
shows that adaptive softness contributes a further 1.85 points over
basic soft thinking, compared with the 0.65-point gain from hard to
basic soft thinking. Thus, the larger benefit from adjusting softness
across reasoning steps observed on OmniSpatial persists under zero-shot
transfer.
\begin{wraptable}{r}{0.56\textwidth}
\vspace{-8pt}
\centering
\caption{\small Zero-shot results on MindCube~\cite{wang2026mindcubespatialmentalmodeling}
across three spatial mental-modeling settings. Proprietary models are
included as reference points, and Overall is weighted by setting size.}
\label{tab:mindcube}
\scriptsize
\setlength{\tabcolsep}{2.5pt}
\renewcommand{\arraystretch}{0.94}

\begin{tabularx}{\linewidth}{@{}Xcccc@{}}
\toprule
\textbf{Method} &
\textbf{Overall} &
\textbf{Rotation} &
\textbf{Among} &
\textbf{Around} \\
\midrule

\multicolumn{5}{@{}l}{\textbf{Reference Baseline}} \\
\reftext{Random Choice}
    & \reftext{32.35}
    & \reftext{36.36}
    & \reftext{32.29}
    & \reftext{30.66} \\

\midrule
\multicolumn{5}{@{}l}{\textbf{Proprietary Models}} \\
\closedtext{Gemini-2.5-Pro} \closedtext{\cite{team2023gemini}}
    & \closedtext{47.05}
    & \closedtext{85.50}
    & \closedtext{25.95}
    & \closedtext{38.40} \\
\closedtext{Claude-4-Sonnet} \closedtext{\cite{anthropic2024model}}
    & \closedtext{44.75}
    & \closedtext{48.42}
    & \closedtext{44.21}
    & \closedtext{47.62} \\

\midrule
\multicolumn{5}{@{}l}{\textbf{Open-weight Models}} \\
InternVL3-8B \cite{wang2025internvl3}
    & 37.50& 26.00 & \textbf{42.03} & \textbf{36.00} \\
Qwen3-VL-8B-Thinking (Base) \cite{bai2025qwen3}
    & 33.62 & 45.14 & 33.26 & 30.44 \\

\midrule
\multicolumn{5}{@{}l}{\textbf{Soft-Thinking Models}} \\
LVR \cite{li2026latent}
    & 29.53 & 36.07 & 29.44 & 26.59 \\
Laser \cite{wang2026forest}
    & 27.29 & 37.65 & 30.12 & 21.02 \\

\midrule
\multicolumn{5}{@{}l}{\textbf{Spatial Reasoning Models}} \\
SpaceMantis \cite{chen2024spatialvlm}
    & 22.81 & 37.65 & 21.26 & 29.32 \\
SpaceQwen \cite{chen2024spatialvlm}
    & 33.28 & 38.02 & 33.71 & 26.32 \\

\midrule
\multicolumn{5}{@{}l}{\textbf{Proposed Method}} \\
\textbf{Soft Spatial Reasoning (Ours)}
    & \textbf{38.13}& \textbf{47.36} & 38.18 & 32.37 \\

\bottomrule
\end{tabularx}
\vspace{-4mm}
\end{wraptable}
Unlike on OmniSpatial, adaptive softness improves over basic soft thinking
in every category. The gains are nevertheless uneven: Relative Position
(+3.30) and Spatial Navigation (+3.37) improve substantially more than
Orientation (+0.49) and Size \& Scale (+0.79). The consistent direction
of these improvements supports the transferability of learned softness
control, while their differing magnitudes indicate that its benefit
depends on the spatial task.

\noindent\textbf{What Adaptive Softness Adds to Spatial Mental Modeling.}
MindCube~\cite{wang2026mindcubespatialmentalmodeling} requires integrating
limited views to infer unseen spatial relations and reason about
hypothetical movements. In this zero-shot setting, Soft Spatial Reasoning
achieves 38.13 weighted-average accuracy, the highest among the compared
non-proprietary models, and improves on its backbone by 4.51 points
(\Cref{tab:mindcube}). Improvements span all three settings, extending
the framework's benefits to reasoning about partially observed scenes.
Adaptive softness may support this process by preserving plausible
spatial interpretations while combining evidence from different views.

\begin{wraptable}{r}{0.55\textwidth}
\vspace{-4mm}
\centering
\small
\setlength{\tabcolsep}{4pt}
\caption{\small OmniSpatial \textit{Perspective Taking}
ablations~\cite{jia2025omnispatial}. All variants share the backbone
and policy GRPO setup, with post-training and evaluation on the
corresponding training and test splits. Averages are sample-weighted.}
\label{tab:ablation}
\resizebox{\linewidth}{!}{
\begin{tabular}{lcccc}
\toprule
Variant & Avg. & \shortstack[c]{Ego\\centric} &
\shortstack[c]{Allo\\centric} &
\shortstack[c]{Hypo\\thetical} \\
\midrule
Soft Spatial Reasoning (Full)
    & \textbf{42.75} & \textbf{81.37} & \textbf{32.53} & \textbf{41.46} \\
\quad w/o $h_{i,t}$ in AdaptSoft
    & 39.22 & 77.45 & 29.52 & 36.14 \\
\quad w/o predictive uncertainty
    & 40.64 & 78.43 & 30.85 & 38.55 \\
\quad w/o gradient alignment
    & 38.68 & 76.47 & 29.26 & 34.94 \\
\quad w/o gradient centering
    & 41.18 & 79.41 & 31.12 & 39.76 \\
\bottomrule
\end{tabular}}
\vspace{-4mm}
\end{wraptable}

\noindent\textbf{Ablation Study.}
All four ablations reduce accuracy in every perspective-taking category
(\Cref{tab:ablation}). Replacing AdaptSoft's step-specific gradient
alignment with the rollout-level task advantage produces the largest
average decline (4.07 points), including a 6.52-point drop in
\emph{Hypothetical} reasoning. This suggests that the shared GRPO
signal provides less guidance for controlling softness at individual
steps. Removing the final-layer hidden state $h_{i,t}$ from AdaptSoft
reduces average accuracy by 3.53 points, compared with 2.11 points
without predictive uncertainty, supporting the use of the ongoing
reasoning state alongside predictive uncertainty to control softness.
Omitting gradient centering reduces accuracy in all three categories,
with a smaller average decline of 1.57 points.

\section{Conclusion}
We introduced \textbf{Soft Spatial Reasoning}, a post-training framework that enables adaptive soft thinking for spatial reasoning in LVLMs. \textbf{AdaptSoft} controls softness using the reasoning state and predictive uncertainty, with step-specific guidance from gradient alignment. Across three benchmarks, our framework achieves the highest weighted-average accuracy among the evaluated non-proprietary models, with gains extending to unseen benchmarks.

\noindent\textbf{Limitations.}
AdaptSoft lacks an explicit measure of uncertainty in visual evidence. Future work could incorporate visual uncertainty at individual CoT steps, allowing softness to reflect ambiguity in both visual evidence and language generation.

\bibliography{iclr2027_conference}
\bibliographystyle{iclr2027_conference}

\newpage
\appendix

\section{Appendix: Additional Analysis}
\label{app:additional_analysis}
\begin{wrapfigure}{r}{0.55\textwidth}
\vspace{-5mm}
\includegraphics[width=\linewidth]{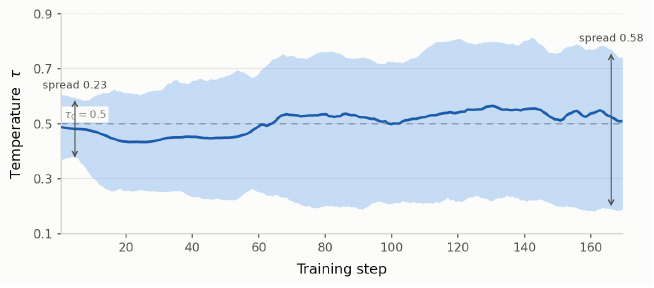}
\caption{\small \textbf{AdaptSoft Temperatures During Training.}
The line shows the mean temperature $\tau_{i,t}$ across soft steps in each rollout
batch; shading spans their minimum and maximum. The dashed line marks
$\tau_0=0.5$.}
\label{fig:figure5}
\end{wrapfigure}
\paragraph{AdaptSoft Learns to Vary Softness.}
As training progresses, \Cref{fig:figure5} shows AdaptSoft assigning
temperatures over a wider range while their mean stays close to
$\tau_0=0.5$ ($0.488$ to $0.510$).
\begin{wrapfigure}{r}{0.55\textwidth}
\vspace{-5mm}
\includegraphics[width=\linewidth]{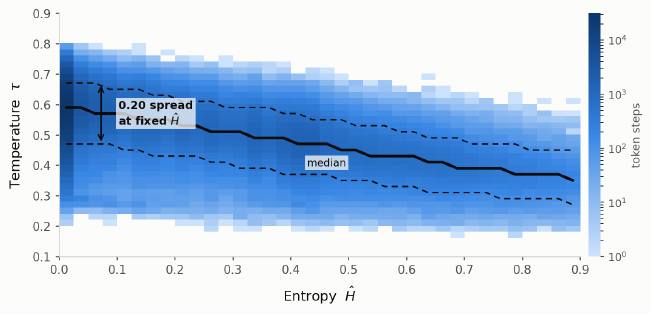}
\caption{\small AdaptSoft Temperature and Predictive Uncertainty.
Temperatures and predictive uncertainty are measured across soft
reasoning steps during inference on the test set. Color shows the
number of steps on a logarithmic scale. The solid line shows median
temperature in each entropy bin; dashed lines mark the 10th and 90th
percentiles.}
\label{fig:figure6}
\end{wrapfigure}
The min--max span grows from
$0.23$ to $0.58$, extending on both sides of the base value. AdaptSoft thus produces broader mixtures at some steps
and more concentrated ones at others without shifting average softness.
This spread develops over successive updates and persists later in
training, revealing learned variation that the stable mean would obscure.
Basic soft thinking can vary mixture weights as token probabilities
change, but its temperature remains fixed. AdaptSoft also learns how
soft each mixture should be; \Cref{fig:figure4} illustrates these
adjustments within an individual soft CoT.

\paragraph{Predictive Uncertainty and Softness.}
\Cref{fig:figure6} plots the temperature and predictive uncertainty
at each soft reasoning step across inference rollouts of the trained
model. As normalized top-$k$ entropy increases, median temperature
falls from $0.590$ to $0.350$, with a correlation of $-0.613$ across
all steps.
\begin{wrapfigure}{r}{0.55\textwidth}
\vspace{-5mm}
\includegraphics[width=\linewidth]{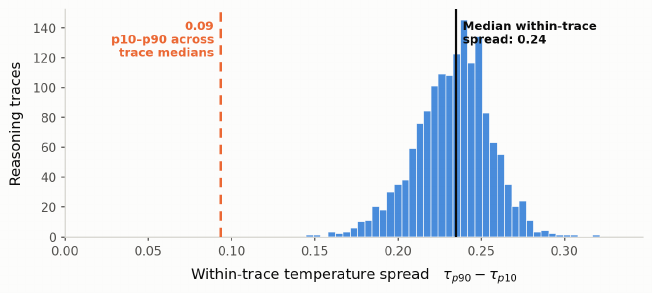}
\caption{\small \textbf{AdaptSoft Varies Softness Within CoTs.}
The histogram shows the 90th--10th percentile temperature spread
within each test-set reasoning trace generated during inference. The
solid line marks the median within-trace spread; the dashed line marks
the corresponding spread across trace-median temperatures.}
\vspace{-5mm}
\label{fig:figure7}
\end{wrapfigure}
When
probability is spread across more candidates, these lower temperatures
concentrate their influence and may reduce interference from conflicting
spatial interpretations. AdaptSoft also assigns substantially different temperatures at similar
uncertainty. Within every entropy bin, the 10th--90th percentile range
is $0.18$--$0.20$, nearly as large as the $0.24$ change in median across
the full entropy range. Even the lowest-entropy bin, containing $48.4\%$
of steps, spans $0.470$--$0.670$ between these percentiles. Removing the hidden-state
input leaves AdaptSoft with predictive uncertainty alone and lowers
average accuracy by $3.53$ points (\Cref{tab:ablation}).

\paragraph{AdaptSoft Varies Softness Within CoTs.}
\Cref{fig:figure7} compares temperature variation within individual
test-set CoTs during inference with variation across their median
temperatures. The median within-trace spread is $0.235$, about $2.5$
times the $0.093$ spread across trace medians: AdaptSoft changes
softness more as a CoT unfolds than it differs between CoTs.
Even the smallest within-trace spread is $0.145$, showing that these
changes occur across the test set. Correct and incorrect traces have
similar median spreads ($0.233$ and $0.237$), so the variation is not
limited to successful answers. It also persists in short and long
traces ($0.218$ below 100 steps and $0.241$ at 300 steps or more),
rather than arising simply because longer CoTs offer more steps
over which temperature can change.

\begin{table}[t]
\centering
\caption{\small Post-training and evaluation configuration for Soft Spatial Reasoning.}
\label{tab:hyperparams}
\footnotesize
\setlength{\tabcolsep}{4pt}
\renewcommand{\arraystretch}{0.98}
\begin{tabularx}{\linewidth}{@{}l l X@{}}
\toprule
\textbf{Hyperparameter} & \textbf{Value} & \textbf{Notes} \\
\midrule

\multicolumn{3}{@{}l}{\textbf{\textit{Model configuration}}} \\
Backbone
    & Qwen3-VL-8B-Thinking
    & -- \\
Language-model adaptation
    & Full fine-tuning
    & No LoRA \\
Numerical precision
    & BF16
    & FSDP2 mixed precision \\
Parameter sharding
    & FSDP2
    & Parameter and optimizer offload \\

\midrule
\multicolumn{3}{@{}l}{\textbf{\textit{Data preparation}}} \\
Training-image pixel budget
    & $128\times28^2$
    & Aspect ratio preserved; dimensions floored to multiples of 28 \\
Evaluation image resolution
    & Native
    & No resizing \\
Maximum prompt length
    & 10{,}240 tokens
    & Overlong prompts discarded \\

\midrule
\multicolumn{3}{@{}l}{\textbf{\textit{Training configuration}}} \\
Training GPUs
    & 8
    & Two nodes $\times$ four H100s \\
Training steps
    & 200
    & $\approx2$ epochs \\
Prompt batch size
    & 64
    & Per GRPO step \\
Rollouts per prompt
    & $G=8$
    & GRPO group size \\
Optimizer mini-batch size
    & 32
    & Two prompt mini-batches per GRPO batch \\
Micro-batch size
    & 4 sequences/GPU
    & Dynamic batching disabled \\
Maximum training response length
    & 2{,}048 tokens
    & - \\

\midrule
\multicolumn{3}{@{}l}{\textbf{\textit{Policy optimization}}} \\
Optimizer
    & AdamW
    & Policy parameters \\
Learning rate
    & $1\times10^{-6}$
    & -- \\
Learning-rate schedule
    & Cosine
    & 10\% minimum-LR floor \\
Warmup ratio
    & 0.05
    & -- \\
KL coefficient
    & $1\times10^{-3}$
    & Loss-only; frozen reference policy \\
Entropy coefficient
    & 0.0
    & No entropy bonus \\
Advantage estimation
    & GRPO
    & Group-normalized \\

\midrule
\multicolumn{3}{@{}l}{\textbf{\textit{Soft thinking}}} \\
Candidate set
    & Top-$k=5$
    & Per reasoning step \\
Softness temperature
    & Step-specific
    & Predicted by AdaptSoft \\
Gumbel-noise scale
    & 1.0
    & Added to candidate log-probabilities \\
Application span
    & Reasoning tokens only
    & Ends at \texttt{</think>}; discrete answer \\

\midrule
\multicolumn{3}{@{}l}{\textbf{\textit{AdaptSoft $f_\phi$}}} \\
Projection dimension
    & $d_p=8$
    & Fixed random projection \\
AdaptSoft architecture
    & Two-layer MLP
    & Width 256; GELU \\
Temperature map
    & $\tau_{i,t}=0.5+0.4\tanh(u_{i,t})$
    & $\tau_{i,t}\in(0.1,0.9)$ \\
AdaptSoft initialization
    & $a=0,\ b=0$
    & Starts at $\tau=0.5$ \\
Entropy standardization
    & $\mu_H=0.173,\ \sigma_H=0.224$
    & Estimated from the initial policy and held fixed \\
AdaptSoft optimizer
    & AdamW, $1\times10^{-3}$
    & Separate from policy optimizer \\
Gradient centering
    & Per rollout
    & Across soft reasoning steps \\
Reference subset
    & 25\% of each batch
    & Disjoint and all-gathered \\

\midrule
\multicolumn{3}{@{}l}{\textbf{\textit{Reward}}} \\
Answer-reward weight
    & 1.0
    & Exact answer match \\
Format-reward weight
    & 0.2
    & One \texttt{<think>} block; final letter A--D \\

\midrule
\multicolumn{3}{@{}l}{\textbf{\textit{Evaluation}}} \\
Samples per question
    & 8
    & Mean@8 \\
Decoding temperature
    & 0.6
    & -- \\
Decoding top-$k$
    & 5
    & -- \\
Maximum response length
    & 3{,}072 tokens
    & -- \\

\midrule
\multicolumn{3}{@{}l}{\textbf{\textit{Infrastructure}}} \\
RL framework
    & \texttt{verl} 0.8.0
    & GRPO; FSDP2 \\
Rollout engine
    & SGLang 0.5.12
    & TP${}=1$; DP${}=8$ \\
GPU memory fraction
    & 0.4
    & Rollout engine \\
Orchestration
    & Ray and Slurm
    & Multi-node execution \\

\bottomrule
\end{tabularx}
\end{table}

\section{Appendix: Implementation Details}
\label{sec:rationale}

\subsection{Hyperparameter Settings}
\label{app:implementation_details}

\Cref{tab:hyperparams} reports the main hyperparameters used for Soft Spatial Reasoning post-training, including the GRPO training setup, optimization settings, and reward configuration.

\subsection{Reasoning Format and Discrete Answer Generation}
\label{app:system_prompt}

\begin{figure*}[t]
\centering
\includegraphics[width=1.0\textwidth]{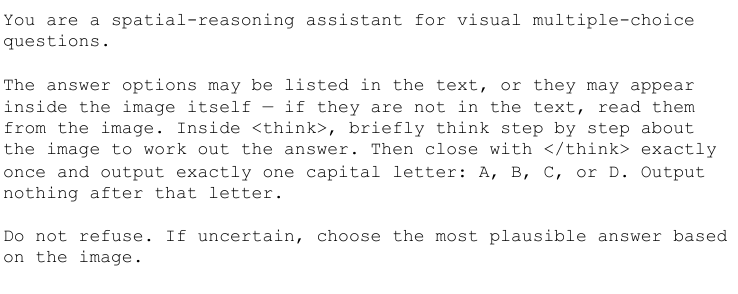}
\caption{\small System prompt used to standardize the reasoning format during Soft Spatial Reasoning training.}
\vspace{-5mm}
\label{fig:figure8}

\end{figure*}

We use the fixed system prompt in \Cref{fig:figure8} during training to enforce
a consistent reasoning format. Since the chat template of Qwen3-VL-Thinking
\cite{bai2025qwen3} automatically prefills the opening \texttt{<think>} token
after the user message, the prompt does not ask the model to generate
\texttt{<think>}; it only requires the model to close the reasoning segment with
\texttt{</think>}.

During rollout generation, each reasoning step feeds a mixture of token
embeddings back to the model. We also retain the highest-weight token
at each step as a discrete \emph{spine}, which marks where reasoning
ends. When the spine emits \texttt{</think>}, subsequent steps sample
discrete tokens from the full vocabulary to generate the final answer.
The delimiter itself remains part of the soft reasoning phase; the
switch occurs immediately afterward.

\section{Appendix: Technical Derivations}
\subsection{Derivation of the Gumbel-Reparameterized Likelihood}
\label{app:gumbel_likelihood}

The likelihood in Equation~\ref{eq:current_gumbel_likelihood} follows the
Gumbel-reparameterized construction of
SofT-GRPO~\cite{zheng2025soft}. The sampled variable at each soft reasoning
step is the perturbed score vector $z_{i,t}$, while the soft state is
constructed deterministically from $z_{i,t}$ and $\tau_{i,t}$. Conditioning
on $(I,q,s_{i,<t})$ is omitted below for compactness.

During rollout, each score is obtained by adding independent standard Gumbel
noise to the rollout-policy log-probability. Under the current policy, the
same recorded score implies a different noise value:
\begin{equation}
\begin{aligned}
z_{i,t,k}
&=
\log\bar{p}_{i,t,k}+\gamma_{i,t,k},
&
\gamma_{i,t,k}
&\overset{\mathrm{i.i.d.}}{\sim}\operatorname{Gumbel}(0,1),
\\
f_{\mathrm{Gum}}(\gamma)
&=
\exp\!\left[-\gamma-\exp(-\gamma)\right],
&
\widetilde{\gamma}_{i,t,k}^{\,\theta}
&=
z_{i,t,k}-\log\bar{p}_{i,t,k}^{\,\theta}.
\end{aligned}
\end{equation}
The transformation from $\widetilde{\gamma}_{i,t,k}^{\,\theta}$ to
$z_{i,t,k}$ is additive and therefore has unit Jacobian. Independence
across vocabulary tokens gives the density and its logarithm, recovering
Equation~\ref{eq:current_gumbel_likelihood}:
\begin{equation}
P_\theta(z_{i,t})
=\prod_{k=1}^{|\mathcal{V}|}
f_{\mathrm{Gum}}\!\left(\widetilde{\gamma}_{i,t,k}^{\,\theta}\right),
\qquad
\log P_\theta(z_{i,t})
=\sum_{k=1}^{|\mathcal{V}|}
\left[-\widetilde{\gamma}_{i,t,k}^{\,\theta}
-\exp\!\left(-\widetilde{\gamma}_{i,t,k}^{\,\theta}\right)\right].
\end{equation}

Under the rollout policy, the implied noise equals the originally sampled
value. Evaluating the same score vector under both policies therefore gives
\begin{equation}
\begin{gathered}
\widetilde{\gamma}_{i,t,k}^{\,\theta_{\mathrm{old}}}
=
z_{i,t,k}-\log\bar{p}_{i,t,k}
=
\gamma_{i,t,k},
\qquad
\log P_{\theta_{\mathrm{old}}}(z_{i,t})
=
\sum_{k=1}^{|\mathcal{V}|}
\left[-\gamma_{i,t,k}-\exp(-\gamma_{i,t,k})\right],
\\
\rho_{i,t}^{\mathrm{soft}}(\theta)
=
\frac{P_\theta(z_{i,t})}
     {P_{\theta_{\mathrm{old}}}(z_{i,t})}
=
\exp\!\left(
\log P_\theta(z_{i,t})
-
\log P_{\theta_{\mathrm{old}}}(z_{i,t})
\right).
\end{gathered}
\end{equation}
The likelihood ratio in Equation~\ref{eq:soft_likelihood_ratio}
evaluates the recorded perturbed scores $z_{i,t}$ under both policies.
Equation~\ref{eq:gumbel_soft_state} then constructs $s_{i,t}$
deterministically from $(z_{i,t},\tau_{i,t})$, so the ratio requires
no separate density over soft states.

\subsection{Stop-Gradient Reconstruction for AdaptSoft}
\label{app:adaptsoft_stopgrad}

Training AdaptSoft requires gradients through the step-specific temperature,
but retaining the language-backbone computation graph for every rollout step
would be memory intensive. We therefore store the projected hidden state,
standardized entropy, and AdaptSoft output during rollout. Let
\begin{equation}
x_{i,t}^{\mathrm{roll}}
=
\left[
\mathbf{P}\,\operatorname{LN}(h_{i,t})
\;\middle\Vert\;
\widehat{H}_{i,t}
\right],
\qquad
u_{i,t}^{\mathrm{roll}}
=
f_{\phi_{\mathrm{roll}}}
\!\left(x_{i,t}^{\mathrm{roll}}\right)
\label{eq:stored_adaptsoft}
\end{equation}
denote the stored input and output, where $\phi_{\mathrm{roll}}$ denotes the
AdaptSoft parameters used to generate the rollout.

During the update, AdaptSoft recomputes its output from the stored input
using the current parameters $\phi$. The operator $\operatorname{sg}(\cdot)$
stops gradients through recorded values; the construction below preserves
the rollout temperature in the forward pass while allowing gradients to
reach $\phi$:
\begin{equation}
\begin{gathered}
\widetilde{u}_{i,t}
= f_\phi\!\left(\operatorname{sg}(x_{i,t}^{\mathrm{roll}})\right),\\
\begin{aligned}
u_{i,t}^{\mathrm{upd}}
&= \operatorname{sg}(u_{i,t}^{\mathrm{roll}})
   +\widetilde{u}_{i,t}
   -\operatorname{sg}(\widetilde{u}_{i,t}),
\qquad
\tau_{i,t}^{\mathrm{upd}}
= \tau_0+\Delta\tanh(u_{i,t}^{\mathrm{upd}}),\\
u_{i,t}^{\mathrm{upd}}
&= u_{i,t}^{\mathrm{roll}},
\qquad
\tau_{i,t}^{\mathrm{upd}}
= \tau_{i,t}^{\mathrm{roll}}
\quad\text{(forward pass)}.
\end{aligned}
\end{gathered}
\label{eq:adaptsoft_stopgrad}
\end{equation}
With the recorded perturbed scores, the forward pass reproduces the
rollout mixture weights and, if token embeddings are unchanged, the
rollout soft state. Gradients to $\phi$ pass through the recomputed
AdaptSoft output:
\begin{equation}
\nabla_{\phi}u_{i,t}^{\mathrm{upd}}
=
\nabla_{\phi}\widetilde{u}_{i,t},
\qquad
\nabla_{\phi}\tau_{i,t}^{\mathrm{upd}}
=
\Delta
\left[1-\tanh^2\!\left(u_{i,t}^{\mathrm{roll}}\right)\right]
\nabla_{\phi}\widetilde{u}_{i,t}.
\label{eq:adaptsoft_backward}
\end{equation}
Recomputing AdaptSoft from its stored, detached input gives the update
a gradient path to $\phi$ without another backbone pass to obtain
$h_{i,t}$ or backpropagation through that stored state.

\subsection{Differentiable Reconstruction and AdaptSoft Gradient Propagation}
\label{app:adaptsoft_gradient}

AdaptSoft must preserve the temperatures used to generate each rollout while
retaining a gradient path to its current parameters. Let
$\operatorname{sg}(\cdot)$ denote stop-gradient, and define the recorded
AdaptSoft input as
$\xi_{i,t}=
[\mathbf{P}\operatorname{LN}(h_{i,t})
\Vert\widehat{H}_{i,t}]$.
During rollout, AdaptSoft records $\xi_{i,t}$ and its output
$u_{i,t}^{\mathrm{roll}}$, produced using the rollout parameters
$\phi_{\mathrm{roll}}$. At update time, the output is recomputed using the
current parameters and combined with its recorded value:
\begin{equation}
\begin{gathered}
\widetilde{u}_{i,t}
= f_\phi\!\left(\operatorname{sg}(\xi_{i,t})\right),
\qquad
u_{i,t}^{\mathrm{upd}}
= \operatorname{sg}\!\left(u_{i,t}^{\mathrm{roll}}\right)
  +\widetilde{u}_{i,t}
  -\operatorname{sg}\!\left(\widetilde{u}_{i,t}\right),
\\
\tau_{i,t}^{\mathrm{upd}}
= \tau_0+\Delta\tanh\!\left(u_{i,t}^{\mathrm{upd}}\right).
\end{gathered}
\label{eq:adaptsoft_reconstruction}
\end{equation}
In the forward pass,
$\widetilde{u}_{i,t}
-\operatorname{sg}(\widetilde{u}_{i,t})=0$; hence
$u_{i,t}^{\mathrm{upd}}=u_{i,t}^{\mathrm{roll}}$ and
$\tau_{i,t}^{\mathrm{upd}}=\tau_{i,t}^{\mathrm{roll}}$.
The recorded perturbed scores and candidate token identities consequently
reproduce the rollout temperature and mixture weights even if $\phi$ has
changed since rollout generation.

The stop-gradient terms vanish in the backward pass, allowing the recomputed
output to carry gradients to $\phi$. Together with the dependence of the
mixture weights on temperature, this gives
\begin{equation}
\begin{aligned}
\frac{\partial\tau_{i,t}^{\mathrm{upd}}}{\partial\phi}
&=
\Delta
\left[
1-\tanh^2\!\left(
u_{i,t}^{\mathrm{roll}}
\right)
\right]
\frac{\partial\widetilde{u}_{i,t}}{\partial\phi},
\\
\frac{\partial p_{i,t,k}}{\partial\tau_{i,t}}
&=
-\frac{p_{i,t,k}}{\tau_{i,t}^{2}}
\left(
z_{i,t,k}
-
\sum_{j=1}^{|\mathcal{V}|}
p_{i,t,j}z_{i,t,j}
\right).
\end{aligned}
\label{eq:adaptsoft_temperature_gradient}
\end{equation}
Gradients can therefore pass through the mixture weights, the resulting soft
states, and the subsequent language-model computation to the step-specific
temperatures and AdaptSoft parameters. Equation~\ref{eq:adaptsoft_reconstruction} uses the recorded rollout
temperatures in the forward pass and the recomputed AdaptSoft output
to update $\phi$.

\paragraph{Differentiable Gradient Alignment.}
Let $v_{i,t}$ denote the hidden representation provided to the language-model
output layer, with logits $\ell_{i,t}=Wv_{i,t}$. The GRPO logit residual is
detached when constructing the step-specific output-layer gradient. The
reference gradient is similarly detached after aggregation over the reference steps $\mathcal{R}$ included in
$\mathcal{L}_{\mathrm{GRPO}}^{\mathrm{ref}}$:
\begin{equation}
\begin{gathered}
\begin{aligned}
d_{i,t}
&=
\operatorname{sg}\!\left(
\frac{\partial\mathcal{L}_{\mathrm{GRPO}}}
     {\partial\ell_{i,t}}
\right),
&
g_{i,t}
&=
d_{i,t}v_{i,t}^{\top},
\\
G_{\mathrm{ref}}
&=
\operatorname{sg}\!\left(
\sum_{(j,s)\in\mathcal{R}}
d_{j,s}v_{j,s}^{\top}
\right),
&
\alpha_{i,t}
&=
\left\langle g_{i,t},G_{\mathrm{ref}}\right\rangle_F
=
d_{i,t}^{\top}G_{\mathrm{ref}}v_{i,t},
\end{aligned}
\\
\frac{\partial \alpha_{i,t}}{\partial v_{i,t}}
=
G_{\mathrm{ref}}^{\top}d_{i,t}.
\end{gathered}
\label{eq:alignment_outer_product}
\end{equation}
With $d_{i,t}$ and $G_{\mathrm{ref}}$ detached, the alignment score
provides a first-order gradient through $v_{i,t}$ to earlier
temperatures, since $v_{i,t}$ precedes $s_{i,t}$.

\paragraph{Gradient Scaling and Centering.}
We scale the alignment objective using a detached moving estimate of the
root-mean-square alignment score. For the optimization steps $\mathcal{T}$
in each micro-batch, the per-worker values of $q_b$ are averaged before
updating $m_b$:
\begin{equation}
\begin{aligned}
q_b
&=
\sqrt{
\frac{1}{|\mathcal{T}|}
\sum_{(i,t)\in\mathcal{T}}
\operatorname{sg}(\alpha_{i,t})^2
},
&
m_b
&=
\beta m_{b-1}+(1-\beta)q_b,
\\
\kappa_b
&=
\min\!\left(
\kappa_{\max},
\frac{c_{\mathrm{sc}}}{m_b+\varepsilon_{\mathrm{sc}}}
\right),
&
\mathcal{L}_{\mathrm{AS}}^{\mathrm{scaled}}
&=
\frac{\kappa_b}{|\mathcal{T}|}
\mathcal{L}_{\mathrm{AS}}
=
-\frac{\kappa_b}{|\mathcal{T}|}
\sum_{(i,t)\in\mathcal{T}}\alpha_{i,t}.
\end{aligned}
\end{equation}
With $\beta=0.99$, $\kappa_{\max}=10^6$,
$c_{\mathrm{sc}}=10^{-3}$, and
$\varepsilon_{\mathrm{sc}}=10^{-30}$, the resulting temperature
gradients are centered within each rollout and applied to AdaptSoft
through a surrogate objective:

\begin{equation}
\begin{gathered}
\begin{aligned}
g_{i,t}^{\tau}
&=
\frac{\partial\mathcal{L}_{\mathrm{AS}}^{\mathrm{scaled}}}
     {\partial\tau_{i,t}},
&
\overline{g}_i^{\tau}
&=
\frac{1}{|\mathcal{T}_i|}
\sum_{t\in\mathcal{T}_i}g_{i,t}^{\tau},
\\
\widetilde{g}_{i,t}^{\tau}
&=
\operatorname{sg}\!\left(
g_{i,t}^{\tau}-\overline{g}_i^{\tau}
\right),
&
\widetilde{\mathcal{L}}_{\mathrm{AS}}(\phi)
&=
\sum_i\sum_{t\in\mathcal{T}_i}
\widetilde{g}_{i,t}^{\tau}\tau_{i,t}^{\mathrm{upd}},
\end{aligned}
\\
\nabla_\phi\widetilde{\mathcal{L}}_{\mathrm{AS}}
=
\sum_i\sum_{t\in\mathcal{T}_i}
\widetilde{g}_{i,t}^{\tau}
\frac{\partial\tau_{i,t}^{\mathrm{upd}}}{\partial\phi}.
\end{gathered}
\end{equation}
Centering removes the component shared by the temperature-gradient
signals within a rollout, retaining their differences across steps. The reference-gradient factors are shared across workers, and AdaptSoft
gradients are averaged before its optimizer step. Only the centered
alignment gradient updates $\phi$: accumulated GRPO gradients on $\phi$
are discarded, and the alignment computation does not accumulate
gradients on $\theta$.

\section{Appendix: Benchmark and Evaluation Details}

\subsection{Benchmarks}
\label{app:benchmark_details}
We evaluate Soft Spatial Reasoning on OmniSpatial~\cite{jia2025omnispatial},
SpatiaLab~\cite{wasi2026spatialab}, and MindCube~\cite{wang2026mindcubespatialmentalmodeling}.
OmniSpatial provides the post-training and held-out evaluation splits; the
other two benchmarks are used only for zero-shot evaluation.

\paragraph{OmniSpatial~\cite{jia2025omnispatial}.}
OmniSpatial comprises more than 8.4K question--answer pairs spanning dynamic
reasoning, spatial interaction, complex spatial logic, and perspective taking.
These four dimensions contain 50 fine-grained task subcategories. We use its
official 6,902-sample training split for post-training and its 1,533-sample
held-out test split for evaluation.

\paragraph{SpatiaLab~\cite{wasi2026spatialab}.}
SpatiaLab comprises 1,400 question--answer pairs drawn from realistic,
unconstrained scenes. Its six categories are relative positioning, depth and
occlusion, orientation, size and scale, spatial navigation, and 3D geometry;
each contains five task types. We evaluate in the multiple-choice setting.
No SpatiaLab samples are used for post-training.

\paragraph{MindCube~\cite{wang2026mindcubespatialmentalmodeling}.}
MindCube contains 21,154 questions across 3,268 images and examines spatial
mental modeling from limited views. Its Rotation, Among, and Around settings
test reasoning about spatial relationships as viewpoints change and objects
become partially visible. We evaluate its multiple-choice questions without
using MindCube samples for post-training.

\subsection{Evaluation protocol}
For all three benchmarks, a response is correct when its selected option
matches the ground-truth answer. We report accuracy by category for
OmniSpatial and SpatiaLab and by setting for MindCube. Overall accuracy for
Soft Spatial Reasoning is computed over all evaluated questions, equivalently weighting each
category or setting by its sample count.

\end{document}